\pdfoutput=1
\documentclass{article}
\PassOptionsToPackage{table}{xcolor}
\usepackage[T1]{fontenc}
\usepackage[utf8]{inputenc}
\usepackage{iclr2027_conference,times}

\usepackage{amsmath,amsfonts,bm}

\def\eqref#1{equation~\ref{#1}}

\def\1{\bm{1}}

\DeclareMathAlphabet{\mathsfit}{\encodingdefault}{\sfdefault}{m}{sl}
\SetMathAlphabet{\mathsfit}{bold}{\encodingdefault}{\sfdefault}{bx}{n}

\usepackage{amssymb}
\usepackage{graphicx}
\usepackage{subcaption}
\usepackage{url}
\usepackage{hyperref}
\usepackage{enumitem}
\usepackage{cuted}
\usepackage{float}
\usepackage{placeins}
\usepackage{booktabs}
\usepackage{tabularx}
\usepackage{needspace}
\usepackage[table]{xcolor}
\usepackage{multirow}
\definecolor{qasrow}{RGB}{235,234,255}
\hypersetup{hidelinks}
\newif\ifmarkchanges
\markchangesfalse
\ifdefined\markedversion\markchangestrue\fi
\newcommand{\revised}[1]{{\ifmarkchanges\color{blue}\else\color{black}\fi#1}}
\newenvironment{revision}{\ifmarkchanges\color{blue}\else\color{black}\fi}{}

\title{QASparse: Query-Adaptive Sparse Attention with Tailored Budgets and Tokens}
\author{%
\textbf{Zhiyuan Chen$^{1}$, Zihao Yang$^{2}$, Quanjiang Li$^{3}$,
Haijiang Li$^{4}$, Jiancheng Pan$^{5}$,}\\
\textbf{Qichao Zhao$^{5}$, Zihan Dong$^{6}$, Qiushi Shi$^{7\dagger}$, Jinshi Liu$^{8\dagger}$}\\
\normalfont $^{1}$Beijing Institute of Technology \quad $^{2}$Uppsala University\\
\normalfont $^{3}$National University of Defense Technology \quad $^{4}$Nanjing University of Science and Technology\\
\normalfont $^{5}$Tsinghua University \quad $^{6}$Georgia Institute of Technology\\
\normalfont $^{7}$Hohai University \quad $^{8}$Shenzhen University\\
\normalfont\texttt{\{chiking0718, shiqiushi001, fearnneeley69\}@gmail.com}
}
\iclrfinalcopy

\begin{document}
\maketitle
\begingroup
\renewcommand\thefootnote{$\dagger$}\footnotetext{Corresponding authors.}
\endgroup
\lhead{Preprint}

\begin{abstract}
Long-sequence inference in large language models is constrained by the high cost of prefilling, where dense attention scales quadratically with context length. Exploiting similarity across attention heads offers a way to reduce redundant computation. Prior approaches in this direction have relied on localized query observations to determine head-level computation reuse or budget allocation. However, inter-head similarity does not imply uniform query requirements, leaving adaptation to varying budget needs and selection preferences across query regions a key challenge. We introduce $\textit{\textbf{\textit{QASparse}}, a \textbf{\textit{Q}}uery-\textbf{\textit{A}}daptive \textbf{\textit{Sparse}} attention mechanism}$ with hardware-efficient execution. QASparse preserves the efficiency of shared cross-head importance estimation while adapting computational budgets and token selections to query requirements, and it executes these dynamic sparse decisions through dedicated kernels. Its query adaptivity stems from two complementary innovations: Local Budget Regression models local variations in computational demand along query positions to estimate budgets for different query regions and attention heads; Boundary Voting lets queries within each head vote over candidates near the selection cutoff using actual query/key scores, refining the retained set while preserving the initial selection size. To reduce the execution overhead of dynamic budgets, we further design length-specialized selection kernels that efficiently support query-adaptive sparse patterns across context lengths. Together, these designs efficiently combine shared estimation with local decisions while adapting both how much attention is computed and which interactions are retained. Experimental results demonstrate improved benchmark scores and faster prefill attention computation compared with existing sparse attention baselines.

\end{abstract}

\section{Introduction}

Large language models (LLMs) are applied to tasks such as long-document understanding and repository-level code processing, motivating efficient long-sequence inference~\citep{AComprehensiveSurveyonLongContextLanguageModeling,huoRepoShapleyShapleyEnhancedContext2026}. However, as input sequences grow longer, dense attention in the prefill stage performs QK computation for all causally visible tokens, incurring quadratic computational cost in sequence length and potentially becoming a major bottleneck in long-context prefilling~\citep{zhaoSurveyLargeLanguage2026,fuChallengesDeployingLongContext2024}. Efficient dense-attention implementations improve execution efficiency through tiling, fusion, and parallelization~\citep{FlashAttention-2:FasterAttentionwithBetterParallelismandWorkPartitioning,youFlashlightPyTorchCompiler2026}, while retaining QK computation that grows quadratically with input length. To reduce this cost, sparse attention performs QK computation only for tokens deemed important, omitting the remaining QK computation~\citep{FlexPrefill:AContext-AwareSparseAttentionMechanismforEfficientLong-SequenceInference,fanFlashPrefillInstantaneousPattern2026,zhangS2OEarlyStopping2026,jonesQUOKAQueryOrientedKV2026a,joTokenSparseAttention2026,songCompactAttention2026}. Effective sparse acceleration therefore requires selecting important tokens with low overhead while largely preserving dense-attention capabilities.

To reduce the importance-estimation overhead of constructing sparse indices, a class of methods compresses queries and keys along the sequence dimension and constructs sparse indices from correlations among the compressed representations~\citep{FlexPrefill:AContext-AwareSparseAttentionMechanismforEfficientLong-SequenceInference,wangSparserBlockSparse2026,liuUnifiedSparseAttention,liuRRAttentionDynamicBlock2026}. Although this approach reduces the number of representations involved in estimation, it may obscure fine-grained correlations and distort block-importance estimates. Another line of work exploits similarities across attention heads to reduce redundant computation by sharing attention scores or sparse patterns~\citep{wangProxyAttnGuidedSparse2025,agarwalCHAIClusteredHead2024,pengAcceleratingPrefillingLongContext2025}. ProxyAttn~\citep{wangProxyAttnGuidedSparse2025} analyzes the similarity among different attention heads and uses pooled representative heads to provide shared importance rankings, together with online head-level budget estimation for sparse index construction. However, relying on a single shared ranking and budgets derived from local Q observations cannot accommodate the varying budget demands across query positions or the head-specific token preferences of individual queries. Consequently, queries may receive excessive or insufficient budgets and fail to select the most relevant tokens, due to the lack of query adaptivity.

To this end, we propose $\textit{\textbf{\textit{QASparse}}, a \textbf{\textit{Q}}uery-\textbf{\textit{A}}daptive \textbf{\textit{Sparse}} attention mechanism}$ with hardware-efficient execution, which adaptively allocates computational budgets and selects KV tokens for different query regions and attention heads. QASparse achieves this query adaptivity through two complementary mechanisms: (1) Local Budget Regression uses attention statistics from sampled queries to model local budget variation, estimating suitable budgets across query regions and attention heads; (2) Boundary Voting uses the predicted budget to locate the selection boundary, then computes QK scores for candidates near that boundary to finalize the retained set. Together, these mechanisms enable QASparse to assign more suitable computational budgets to different query regions while maintaining low selection overhead by restricting additional decisions to a small number of boundary candidates. Furthermore, to reduce the execution overhead of dynamic token-selection decisions, we design Length-Specialized Kernels with dedicated execution paths tailored to different context lengths, thereby efficiently supporting query-adaptive sparse patterns across sequence lengths.

We evaluate QASparse on Llama3.1-8B-Instruct~\citep{grattafioriLlama3Herd2024}, MegaBeam-Mistral-7B-512k~\citep{wuScalingContextNot2025}, GLM-4-9B-Chat-1M~\citep{teamglmChatGLMFamilyLarge2024}, and Qwen2.5-7B-Instruct-1M~\citep{yangQwen251MTechnicalReport2025}, using RULER~\citep{hsiehRULERWhatsReal2024} and InfiniteBench~\citep{zhangInfiniteBench2024} to assess performance across controlled context lengths and long-range dependencies in very long contexts, respectively. Experimental results show that, compared with existing sparse-attention baselines, QASparse improves benchmark performance while accelerating attention computation during the prefill stage. Our method achieves a 2.81$\times$ speedup over FlashAttention-2 at 128K context length.

\section{Sparse Attention: Observations and Motivation}
\label{sec:observations}

\paragraph{Sparse Attention Balances Fidelity and Efficiency}
Sparse attention reduces computational cost by omitting substantial QK computation while preserving dense-attention performance. For query $q$ in head $h$, let $S_{h,q}$ denote the selected key tokens, and let $x_{h,q}$ and $y_{h,q}(S_{h,q})$ denote the dense and sparse attention outputs, respectively. Sparse selection can be formulated as
\begin{equation}
\label{eq:sparse-selection-goal}
\min_{S}\sum_{h,q}
\left[
\left\|x_{h,q}-y_{h,q}(S_{h,q})\right\|_2^2
+\lambda |S_{h,q}|
\right],
\end{equation}
which balances output approximation against QK computation.
However, QK computation alone does not determine latency, which also depends on GPU parallelism and workload balance. Query-Adaptive sparse attention must therefore jointly optimize algorithmic and execution efficiency while preserving selection semantics.
\begin{figure*}[t]
  \centering
  \captionsetup[subfigure]{font=small,labelformat=parens,labelsep=space,justification=centering,singlelinecheck=true,skip=2pt}
  \begin{subfigure}[t]{0.32\textwidth}
    \centering
    \includegraphics[width=\linewidth]{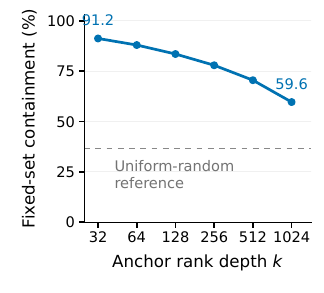}
    \caption{Cross-head sharing}
    \label{fig:cross-head-agreement}
  \end{subfigure}\hfill%
  \begin{subfigure}[t]{0.32\textwidth}
    \centering
    \includegraphics[width=\linewidth]{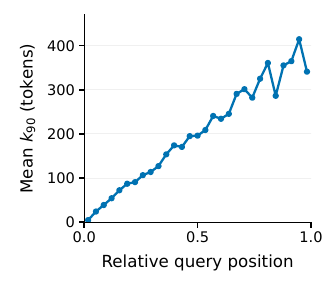}
    \caption{Budget variation}
    \label{fig:query-local-budget}
  \end{subfigure}\hfill%
  \begin{subfigure}[t]{0.32\textwidth}
    \centering
    \includegraphics[width=\linewidth]{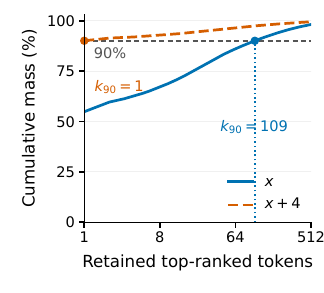}
    \caption{Local budget drop}
    \label{fig:sink-budget-drop}
  \end{subfigure}
  \caption{\textbf{(a)} Cross-head Top-$k$ overlap decreases as $k$ increases. \textbf{(b)} The budget for target attention coverage grows overall with query position but exhibits local variation. \textbf{(c)} Nearby queries can require substantially different budgets for the same 90\% coverage.}
  \label{fig:observations}
\end{figure*}

\paragraph{Attention Heads Share Salient Context Positions}
In LLM attention, highly ranked tokens in one head often overlap with important tokens in other heads, although this overlap weakens as the rank range expands (Figure~\ref{fig:observations}(a)). This cross-head sharing enables shared ranking, where pooled Q/K representations provide a coarse ordering reusable across heads at low cost~\citep{wangProxyAttnGuidedSparse2025}. However, shared importance does not imply identical relative importance across heads, so token selection must also account for each head's attention distribution.

\paragraph{Varying Queries Require Adaptive Attention Budgets}
Prior work~\citep{wangProxyAttnGuidedSparse2025} estimates a head-level budget from the final query block and reuses it along the query axis. However, the budget required for the same attention coverage varies with query position, showing both global trends and local fluctuations (Figure~\ref{fig:observations}(b,c)). A fixed head-level budget may therefore over- or under-allocate tokens, motivating local budget estimation along the query axis.

These observations motivate reusing cross-head importance estimates while adapting budgets locally along the query axis and refining candidate selection with head-specific evidence. The next section develops Local Budget Regression and Boundary Voting around these two forms of adaptation.


\section{Query-Adaptive Selection}
\label{sec:query-adaptive-selection}
Our method reuses a shared ranking method~\citep{wangProxyAttnGuidedSparse2025}: We average-pool Q and K vectors across heads at matching positions and compute their attention scores to obtain a shared token ranking. Building on this shared ranking, we propose \textbf{Query-Adaptive Selection}, which consists of three modules: (1) \textbf{Local Budget Regression}, (2) \textbf{Boundary Voting}, and (3) \textbf{Length-Specialized Kernels}.
\subsection{Local Budget Regression}
The budget variation observed in Section~\ref{sec:observations} (Figures~\ref{fig:query-local-budget} and~\ref{fig:sink-budget-drop}) motivates local estimation along the query axis: budget demand can increase overall while exhibiting local dips, and nearby queries can have substantially different requirements.
\paragraph{Representative Query Sampling.} We select a representative query $\widehat q$ every $s$ positions along the query sequence and compute its attention distribution $\widehat a$, normalized over all causally visible key tokens:
\begin{equation}
\label{eq:qas-sampled-attention}
\widehat a=\operatorname{softmax}\!\left(\frac{K\widehat q}{\sqrt d}\right).
\end{equation}
\paragraph{Budget Estimation along the Shared Ranking.} We accumulate the target head's attention weights over visible tokens along the shared ranking $\pi$. The smallest number of tokens required to reach the target coverage $\gamma$ of total attention weight gives the budget observation $b$:
\begin{equation}
\label{eq:qas-budget}
b=\min\left\{m:\sum_{j=1}^{m}\widehat a_{\pi(j)}\ge\gamma\right\}.
\end{equation}
Although the ordering is shared across heads, the attention weights are head-specific, allowing the same ranking to yield different budgets. \revised{For block-level execution of this token-level formulation, 128 consecutive query tokens share the regression budget at their midpoint token.}
\begin{figure*}[t]
  \centering
  \includegraphics[width=\textwidth]{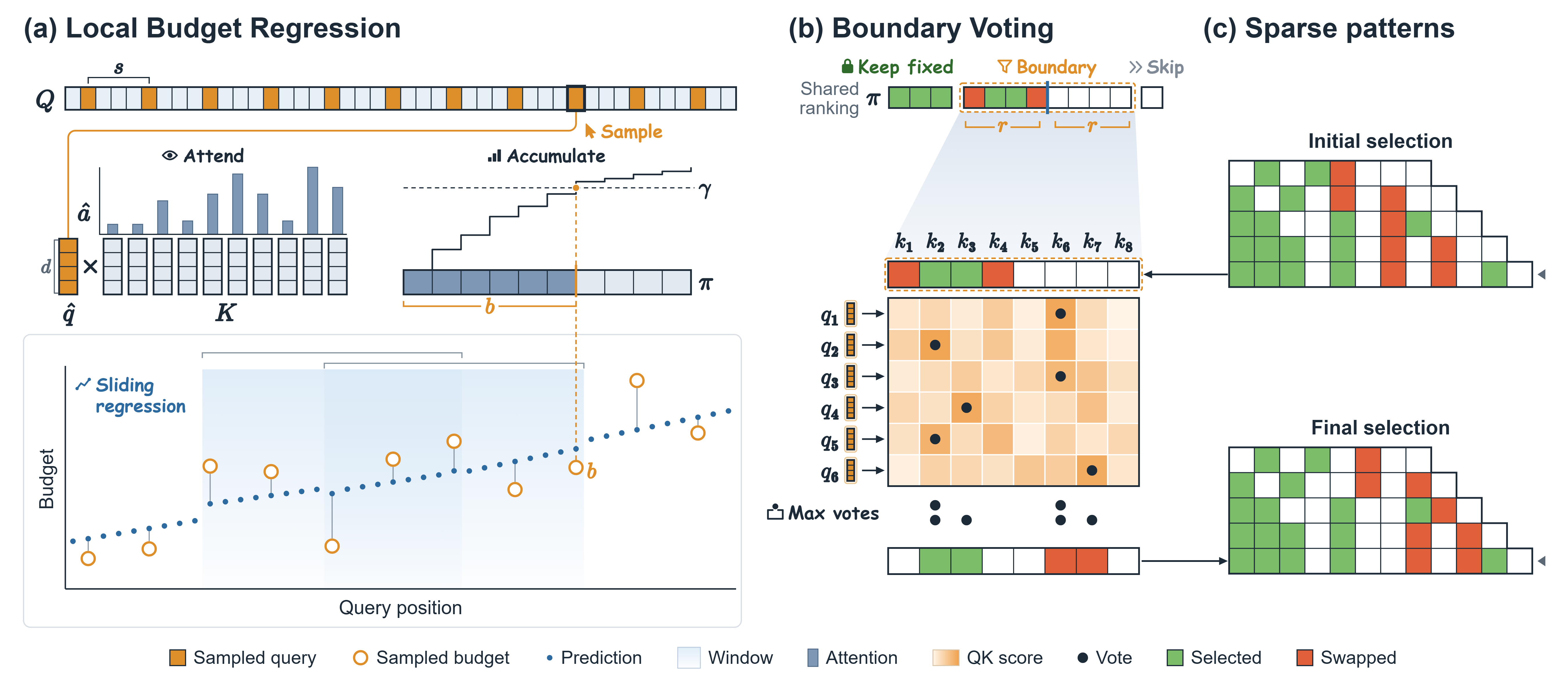}
  \caption{Overview of Query-Adaptive Selection. \textbf{(a)} Budgets are estimated from regularly sampled queries, and sliding-window regression predicts token requirements. \textbf{(b)} \revised{QK scores are used to vote among key blocks near the cutoff of the shared ranking, refining block selection while preserving the voting budget.} \textbf{(c)} Sparse attention patterns before and after Boundary Voting.}
  \label{fig:query-adaptive-selection}
\end{figure*}
\paragraph{Comparing Regression Schemes.} Figure~\ref{fig:sink-budget-drop} in Section~\ref{sec:observations} shows a sharp budget drop between nearby queries. Directly assigning sampled budgets to neighboring queries may propagate the influence of outlying observations to local allocations. We therefore fit linear models to budget trends within neighboring regions and consider three schemes.
\begin{itemize}[leftmargin=1.5em]
\item \textbf{Global Linear Regression.} The overall increase in budget with query position in Figure~\ref{fig:query-local-budget} motivates fitting one line to all sampled budgets. However, a single slope cannot also capture the local drops shown in Figure~\ref{fig:sink-budget-drop}, so we do not adopt this scheme.
\item \textbf{Non-overlapping Piecewise Linear Regression.} We divide sampled positions into disjoint consecutive segments and fit each segment independently. Different regions can have different slopes, but predictions draw on a limited range of observations. In particular, predictions at internal segment boundaries cannot incorporate information from neighboring segments over a broader range. We therefore do not adopt this scheme.
\item \textbf{Sliding-window Linear Regression.} We fit a line to each window of $w$ consecutive sentinel positions and their measured budgets, advance by $\delta$ sentinels, and average all predictions covering each query position. This estimates query-dependent budgets from overlapping local neighborhoods, as illustrated by the sliding regression in Figure~\ref{fig:query-adaptive-selection}.
\end{itemize}
\paragraph{Query-adaptive Budget Assignment.} We assign token budgets to query positions based on the regression predictions and obtain an initial selection from the shared ranking. Boundary Voting then refines its membership while preserving the number of selected tokens.
This adaptive estimation assigns different budgets along the query axis of each head and uses neighborhood information to capture local demand. Sampled-query coverage guides allocation, balancing local adaptation with estimation cost.
\subsection{Boundary Voting}
\revised{Shared ranking provides a low-cost ordering of salient context positions that can be reused across heads (Section~\ref{sec:observations}, Figure~\ref{fig:cross-head-agreement}). Under a fixed block budget, differences across the cutoff can leave a block preferred by the target head outside the retained set. Boundary Voting therefore uses the target head's QK evidence to refine candidate membership on both sides of the shared-ranking cutoff while preserving the number of selected candidate blocks.}

\begin{revision}
Boundary Voting uses $128\times128$ QK blocks so that 128 consecutive Q tokens share the same selected key blocks and are processed jointly on the GPU, reducing repeated key reads while using local token similarity to limit approximation error~\citep{zhang2025spargeattention}.
\paragraph{Boundary Candidate Construction.} We take up to $r$ key blocks on each side of the shared-ranking cutoff to form the candidate set $C$, keeping the higher-ranked block prefix outside this region unchanged.
\paragraph{Head-specific Max Voting.} We sample query and key vectors within each block with stride $t$. Let $T$ contain the sampled queries of the current query block and $K_j$ the sampled keys in candidate block $j$. Each sampled query votes for the candidate block containing its strongest QK interaction:
\begin{equation}
\label{eq:qas-votes}
\begin{aligned}
s_j(q)&=\max_{k\in K_j}q^\top k,\\
v_j&=\sum_{q\in T}\mathbf 1\!\left[j=\operatorname*{arg\,max}_{u\in C}s_u(q)\right].
\end{aligned}
\end{equation}
\paragraph{Budget-preserving Selection.} We rank candidate blocks by their vote counts, retain the same number of boundary blocks as before voting, and combine them with the fixed prefix. This refines the selected block identities under the assigned budget.
\end{revision}
\subsection{Length-Specialized Kernels}
Prior work~\citep{wangProxyAttnGuidedSparse2025,FlexPrefill:AContext-AwareSparseAttentionMechanismforEfficientLong-SequenceInference} often reuses a single attention kernel across context lengths. However, workload size and the amount of schedulable parallel work vary with context length across stages. We therefore design dedicated selection kernels for different context lengths to reduce token-selection overhead. We use \textbf{Budget Estimation along the Shared Ranking}, which computes budgets from sampled queries along the shared ranking, as a representative path.
\begin{figure*}[t]
  \centering
  \includegraphics[width=\textwidth]{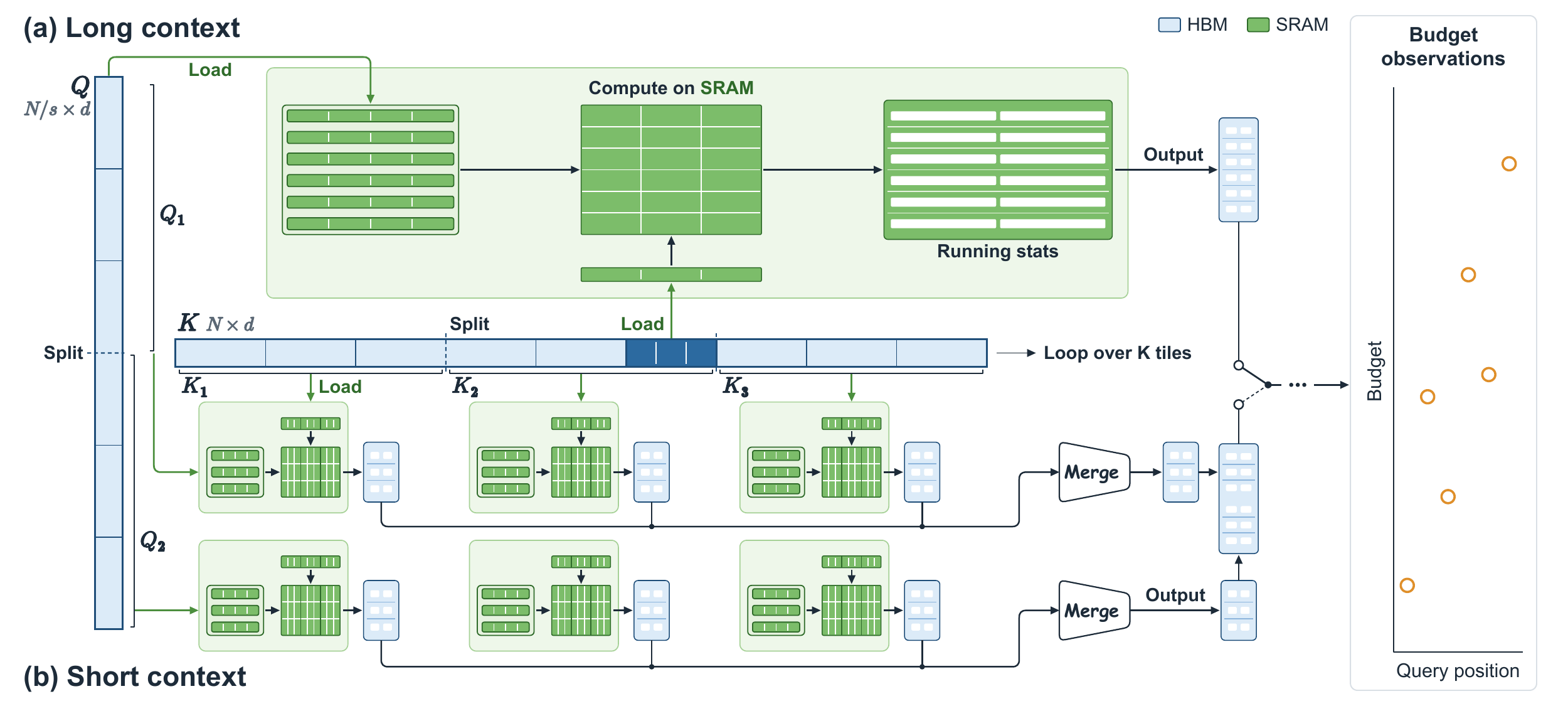}
  \caption{\textbf{(a) Long context:} The kernel keeps a larger, intact group of sampled Q vectors in SRAM and streams through all visible K blocks within one logical workload. \textbf{(b) Short context:} The kernel splits the same Q workload into smaller groups $Q_1$ and $Q_2$ and pairs each group with the K partitions.}
  \label{fig:length-specialized-kernels}
\end{figure*}
\paragraph{Length-specialized workload granularity.} For long contexts, the Budget Estimation in the Shared Ranking stage can load the working sets of multiple sampled queries into fast on-chip storage, keep them resident, and stream visible K vectors to compute against these resident queries.For short contexts, however, an oversized query working set can substantially reduce the number of resulting work units. To increase the number of work units and improve parallelism, we split the sampled-query work that would otherwise be processed together into smaller sampled-query sets and assign them to more independent execution resources.

\paragraph{Context-dependent QK decomposition.} For long contexts, a query group can traverse a continuous visible key range and complete its computation against all visible key vectors as one logical workload. For short contexts, the K range associated with a query group is divided into multiple key-range subproblems so that the work can be distributed across more execution resources and scheduled independently. 

The long-context path uses a reuse-oriented organization to reduce repeated Q/K preparation, whereas the short-context path partitions both query work and K ranges to enlarge the independently schedulable computation. Length-Specialized Kernels thereby improve workload distribution across execution resources and increase GPU parallelism.

\section{Experiments}
\label{sec:experiments}

\subsection{Experimental Setup}
\label{sec:experimental-setup}
In this section, we provide a comprehensive overview of the models and datasets employed in our experiments to assess the performance and capabilities of various long-context language models.
\paragraph{Models}
We use four long-context instruction-following or chat models: Llama-3.1-8B-Instruct~\citep{grattafioriLlama3Herd2024}, Qwen2.5-7B-Instruct-1M~\citep{yangQwen251MTechnicalReport2025}, MegaBeam-Mistral-7B-512k~\citep{wuScalingContextNot2025}, and GLM-4-9B-Chat-1M~\citep{teamglmChatGLMFamilyLarge2024}. MegaBeam-Mistral-7B-512k was further trained from Mistral-7B-Instruct-v0.2. These models span different model families and long-context configurations, allowing us to examine the applicability of the proposed method across architectures and context settings.

\paragraph{Datasets}
We evaluate the models on two benchmarks with different long-context challenges. (i) RULER~\citep{hsiehRULERWhatsReal2024} a synthetic benchmark dataset created to evaluate long-context LLMs with customizable sequence lengths and task complexities. It extends the basic needle-in-a-haystack test as well as introduces new task categories such as multi-hop tracing and aggregation. (ii) InfiniteBench~\citep{zhangInfiniteBench2024} a benchmark dataset designed to test LLMs’ understanding of long dependencies within extensive contexts, with an average token count of 214k. It includes 10 synthetic and real-world tasks across various domains.

\paragraph{Implementation Details}
All experiments are conducted on a single NVIDIA A100 64GB GPU. We implement a custom inference pipeline in PyTorch~\citep{paszkePyTorch2019}, build on FlashAttention~\citep{FlashAttention-2:FasterAttentionwithBetterParallelismandWorkPartitioning} for efficient long-context attention computation, and use Triton~\citep{tilletTriton2019} to optimize GPU computation. We set $\text{block\_size}=128$ in all experiments. We set the cumulative attention threshold to $\gamma=0.95$ for all models and allocate computation budgets adaptively. For Qwen2.5-7B-Instruct-1M, we additionally impose a minimum budget of 2048 tokens.In Local Budget Regression, we sample one representative query at the midpoint of each query block, with an interval of $s=128$ query positions. Each sliding regression window contains $w=10$ consecutive sampled queries and advances by $\delta=2$ sampled queries.In Boundary Voting, we take at most eight candidate key blocks on each side of the shared-ranking cutoff. Voting is organized in $128\times128$ QK blocks, and both query and key vectors are subsampled with stride $t=8$.In the final attention mask, the last two query blocks attend to all causally visible key blocks. For all heads in each layer, we use a shared ranking to guide token selection across all heads.All experiments use greedy decoding to maintain consistency across results.

\paragraph{Baselines}
We compare QASparse with three baselines covering efficient dense attention and adaptive sparse attention:(1) \textbf{FlashAttention}~\citep{FlashAttention-2:FasterAttentionwithBetterParallelismandWorkPartitioning} is an optimized attention implementation that uses hardware-aware computation and efficient memory access to reduce attention computation and memory overhead. We use it as the exact dense-attention baseline for comparing computational savings and model performance.(2) \textbf{FlexPrefill}~\citep{FlexPrefill:AContext-AwareSparseAttentionMechanismforEfficientLong-SequenceInference} dynamically selects between query-aware and vertical-slash sparse patterns according to the input and attention head. We set its pattern-selection threshold to $\tau=0.1$ and its cumulative attention threshold for sparse index selection to $\gamma=0.95$.(3) \textbf{ProxyAttn}~\citep{wangProxyAttnGuidedSparse2025} combines cross-head shared importance rankings with independently estimated head budgets. We configure one proxy head per layer and each head then estimates its own budget with $\gamma=0.95$ and applies that budget to all its query positions, reusing ranking computation while retaining head-specific budgets.

All sparse baselines perform sparse computation during prefill and switch to dense computation during decoding. 
\makeatletter
\@startsection{subsection}{2}{\z@}{-0.8ex}{0.6ex}{\normalsize\sc\raggedright}{Main Results}
\makeatother
\label{sec:main-results}
\begingroup
\fontsize{9.5}{10.5}\selectfont
\setlength{\parskip}{1pt}
\setlength{\intextsep}{2pt}
\captionsetup{font=footnotesize,skip=2pt}

This section presents a comparative analysis of various attention mechanisms. The evaluation includes FlashAttention, FlexPrefill, ProxyAttn, and our proposed method. The comparison aims to demonstrate our approach's efficiency and effectiveness against existing methods.

\vspace{2pt}\noindent\textbf{RULER.}
QASparse achieves the highest mean score among sparse methods on all four models and outperforms both sparse baselines at 128k (Table~\ref{tab:ruler-main}).

\begin{table}[H]
\centering
\fontsize{9}{10.5}\selectfont
\setlength{\tabcolsep}{3pt}
\renewcommand{\arraystretch}{1.1}
\setlength{\aboverulesep}{2pt}
\setlength{\belowrulesep}{2pt}
\caption{Performance comparison of different methods on various models and sequence lengths on RULER. The best results are highlighted in bold.}
\label{tab:ruler-main}
\begin{tabularx}{\linewidth}{ll>{\raggedleft\arraybackslash}X>{\raggedleft\arraybackslash}X>{\raggedleft\arraybackslash}X>{\raggedleft\arraybackslash}X>{\raggedleft\arraybackslash}X>{\raggedleft\arraybackslash}X|>{\raggedleft\arraybackslash}X}
\toprule
\textbf{Models} & \textbf{Methods} & \textbf{4k} & \textbf{8k} & \textbf{16k} & \textbf{32k} & \textbf{64k} & \textbf{128k} & \textbf{Avg} \\
\midrule
\multirow{4}{*}{Llama} & FlashAttention & 95.18 & 93.36 & 93.26 & 89.28 & 85.56 & 77.96 & 89.10 \\
 & FlexPrefill & 94.59 & \textbf{93.53} & 92.58 & 91.59 & 85.18 & 73.67 & 88.52 \\
 & ProxyAttn & 94.12 & 92.25 & 93.35 & 90.34 & 87.24 & 75.15 & 88.74 \\
\rowcolor{qasrow}
 & QASparse & \textbf{94.90} & 93.21 & \textbf{93.75} & \textbf{92.22} & \textbf{87.70} & \textbf{77.82} & \textbf{89.93} \\
\midrule
\multirow{4}{*}{GLM} & FlashAttention & 94.26 & 92.68 & 91.42 & 89.27 & 85.04 & 84.45 & 89.52 \\
 & FlexPrefill & 92.46 & 91.40 & 90.64 & 87.69 & 86.70 & 84.34 & 88.87 \\
 & ProxyAttn & 95.00 & \textbf{93.67} & 91.73 & \textbf{90.57} & \textbf{87.76} & 85.45 & 90.70 \\
\rowcolor{qasrow}
 & QASparse & \textbf{95.05} & 93.46 & \textbf{92.97} & 90.39 & 87.53 & \textbf{86.03} & \textbf{90.91} \\
\midrule
\multirow{4}{*}{Mistral} & FlashAttention & 93.34 & 90.60 & 86.69 & 86.72 & 81.37 & 83.75 & 87.08 \\
 & FlexPrefill & 90.05 & 88.81 & 86.00 & 84.20 & 81.56 & 82.62 & 85.54 \\
 & ProxyAttn & 93.51 & 90.84 & \textbf{88.43} & \textbf{87.09} & \textbf{83.80} & 84.39 & 88.01 \\
\rowcolor{qasrow}
 & QASparse & \textbf{94.21} & \textbf{92.30} & 87.96 & 86.94 & 83.00 & \textbf{85.30} & \textbf{88.29} \\
\midrule
\multirow{4}{*}{Qwen} & FlashAttention & 94.81 & 91.75 & 93.47 & 90.15 & 87.46 & 82.80 & 90.07 \\
 & FlexPrefill & 93.35 & 91.06 & 91.94 & 87.03 & 85.84 & 80.14 & 88.23 \\
 & ProxyAttn & 94.17 & 91.91 & 92.42 & 87.60 & 86.04 & 81.17 & 88.88 \\
\rowcolor{qasrow}
 & QASparse & \textbf{94.35} & \textbf{92.21} & \textbf{92.91} & \textbf{88.95} & \textbf{86.54} & \textbf{82.52} & \textbf{89.58} \\
\bottomrule
\end{tabularx}
\end{table}

\vspace{2pt}\noindent\textbf{InfiniteBench.}
QASparse leads the sparse baselines in mean score on all four models, with strong long-book QA, summarization, and retrieval results (Table~\ref{tab:infinitebench-main}).

\begin{table}[H]
\centering
\fontsize{8}{9.5}\selectfont
\setlength{\tabcolsep}{1.5pt}
\renewcommand{\arraystretch}{1.12}
\setlength{\aboverulesep}{2pt}
\setlength{\belowrulesep}{2pt}
\caption{Performance comparison of different methods on various models and tasks on Infinite Bench. The bold numbers indicate the first highest value in each column.}
\label{tab:infinitebench-main}
\newcommand{\IBtablebody}{%
\toprule
{\fontsize{6.8}{7.8}\selectfont\bfseries Models} & {\fontsize{6.8}{7.8}\selectfont\bfseries Methods} & {\fontsize{6.8}{7.8}\selectfont\bfseries En.Sum} & {\fontsize{6.8}{7.8}\selectfont\bfseries En.QA} & {\fontsize{6.8}{7.8}\selectfont\bfseries En.MC} & {\fontsize{6.8}{7.8}\selectfont\bfseries En.Dia} & {\fontsize{6.8}{7.8}\selectfont\bfseries Zh.QA} & {\fontsize{6.8}{7.8}\selectfont\bfseries Code.Run} & {\fontsize{6.8}{7.8}\selectfont\bfseries Math.Find} & {\fontsize{6.8}{7.8}\selectfont\bfseries Retr.Passkey} & {\fontsize{6.8}{7.8}\selectfont\bfseries Retr.Number} & {\fontsize{6.8}{7.8}\selectfont\bfseries Retr.KV} & {\fontsize{6.8}{7.8}\selectfont\bfseries Avg} \\
\midrule
 & FlashAttention & 35.22 & 34.98 & 81.82 & 20.00 & 42.67 & 0.00 & 16.67 & 96.67 & 100.00 & 60.00 & 48.80 \\
 & FlexPrefill & 32.87 & 28.01 & 72.73 & \textbf{30.00} & 36.40 & 0.00 & 22.22 & 96.67 & 100.00 & 48.00 & 46.69 \\
 & ProxyAttn & 32.09 & 33.22 & 72.73 & 10.00 & \textbf{43.72} & 0.00 & 22.22 & 93.33 & 100.00 & 40.00 & 44.73 \\
\rowcolor{qasrow}
\multirow{-4}{*}{Llama} & QASparse & \textbf{35.13} & \textbf{38.58} & \textbf{72.73} & 10.00 & 41.42 & \textbf{0.00} & \textbf{22.22} & \textbf{96.67} & \textbf{100.00} & \textbf{56.00} & \textbf{47.28} \\
\midrule
 & FlashAttention & 29.32 & 28.57 & 72.73 & 30.00 & 17.10 & 0.00 & 27.78 & 96.67 & 90.00 & 0.00 & 39.22 \\
 & FlexPrefill & \textbf{34.92} & 28.08 & 72.73 & 20.00 & 16.01 & 0.00 & 16.67 & \textbf{100.00} & \textbf{83.33} & 0.00 & 37.17 \\
 & ProxyAttn & 31.13 & 27.49 & 72.73 & 20.00 & \textbf{17.62} & 0.00 & \textbf{27.78} & 96.67 & 80.00 & 0.00 & 37.34 \\
\rowcolor{qasrow}
\multirow{-4}{*}{GLM} & QASparse & 28.78 & \textbf{28.47} & \textbf{72.73} & \textbf{30.00} & 17.52 & \textbf{0.00} & 22.22 & 96.67 & 80.00 & \textbf{0.00} & \textbf{37.64} \\
\midrule
 & FlashAttention & 26.74 & 11.13 & 54.55 & 0.00 & 19.45 & 0.00 & 27.78 & 100.00 & 100.00 & 68.00 & 40.77 \\
 & FlexPrefill & 24.78 & 11.05 & 63.64 & 0.00 & 10.17 & 0.00 & 27.78 & 100.00 & 100.00 & 76.00 & 41.34 \\
 & ProxyAttn & \textbf{26.04} & 10.47 & 54.55 & 0.00 & 14.47 & 0.00 & \textbf{33.33} & 100.00 & 100.00 & 64.00 & 40.29 \\
\rowcolor{qasrow}
\multirow{-4}{*}{Mistral} & QASparse & 24.89 & \textbf{14.19} & \textbf{63.64} & \textbf{0.00} & \textbf{16.54} & \textbf{0.00} & 27.78 & \textbf{100.00} & \textbf{100.00} & \textbf{76.00} & \textbf{42.30} \\
\midrule
 & FlashAttention & 34.68 & 9.66 & 90.91 & 30.00 & 11.32 & 0.00 & 44.44 & 30.00 & 13.33 & 0.00 & 26.43 \\
 & FlexPrefill & 31.35 & \textbf{9.77} & 90.91 & 20.00 & 10.94 & 0.00 & 55.56 & \textbf{26.67} & 10.00 & 0.00 & 25.52 \\
 & ProxyAttn & 31.28 & 8.69 & 81.82 & 20.00 & 13.45 & 0.00 & 44.44 & 10.00 & 6.67 & 0.00 & 21.64 \\
\rowcolor{qasrow}
\multirow{-4}{*}{Qwen} & QASparse & \textbf{38.32} & 8.08 & \textbf{90.91} & \textbf{20.00} & \textbf{13.85} & \textbf{0.00} & \textbf{55.56} & 23.33 & \textbf{13.33} & \textbf{0.00} & \textbf{26.34} \\
\bottomrule
}
\setlength{\tabcolsep}{0pt}
\sbox0{\begin{tabular}{@{}ll*{10}{r}|r@{}}\IBtablebody\end{tabular}}
\setlength{\tabcolsep}{\dimexpr(\linewidth-\wd0)/24\relax}
\begin{tabular}{@{}ll*{10}{r}|r@{}}
\IBtablebody
\end{tabular}
\end{table}

\vspace{2pt}\noindent\textbf{Performance vs. Latency.}
Tuning $\gamma$ balances accuracy and latency. Figure~\ref{fig:pareto} shows improved trade-offs on average and at 4k and 64k; Appendix~\ref{app:pareto-data} reports the full measurements. \revised{At 64k, QASparse achieves up to $1.77\times$ TTFT speedup over FlashAttention-2.}

\begin{figure}[H]
\centering
\includegraphics[width=\linewidth]{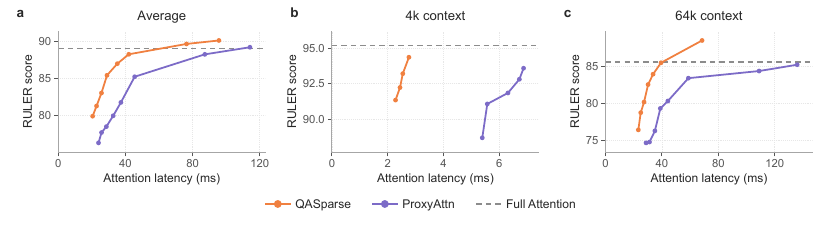}
\caption{RULER quality--latency trade-offs. At 4k, QASparse's specialized short-context kernel supports approximately $3\times$ speedup over ProxyAttn at comparable accuracy. The dashed line denotes Full Attention.}
\label{fig:pareto}
\end{figure}
\endgroup
\FloatBarrier

\Needspace{8\baselineskip}
\subsection{Ablations}\label{sec:ablations}

We examine the effects of kernel specialization, budget regression, shared-ranking count, and Boundary Voting candidate-block count. Appendices~\ref{app:shared-ranking}--\ref{app:budget-regression} report the full results. Additional ablations of full attention on the final two query blocks, budget-label ordering, and voting scores are described in Appendices~\ref{app:final-query-blocks}--\ref{app:vote-scoring}.

\begin{figure}[H]
    \centering
    \includegraphics[width=0.82\linewidth]{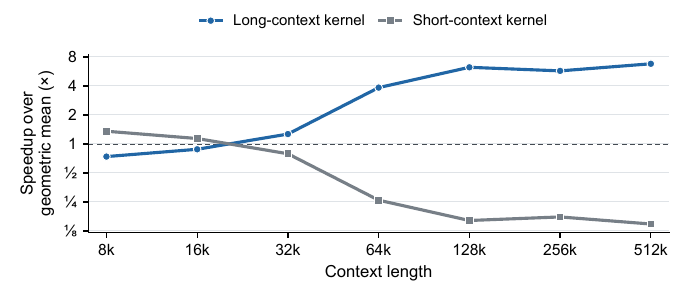}
    \caption{The horizontal $1\times$ line marks the geometric-mean latency of the two kernels at each context length. Speedup is this reference latency divided by the latency of the indicated kernel.}
    \label{fig:kernel-speedup}
\end{figure}

\Needspace{0.40\textheight}
\paragraph{Length-Specialized Kernels.}
We compare the short-context and long-context selection kernels at seven lengths from 8k to 512k. The short-context kernel is faster at 8k and 16k, whereas the long-context kernel is faster at every tested length from 32k onward (Figure~\ref{fig:kernel-speedup}). This crossover supports adapting the work decomposition to context length: finer partitions favor short contexts, while larger work units with greater reuse favor long contexts. Appendix~\ref{app:kernel-timing} reports the selection-kernel latencies.

\paragraph{Budget regression.}
We compare \textbf{Global Linear Regression}, \textbf{Non-overlapping Piecewise Linear Regression}, and \textbf{Sliding-window Linear Regression}, as defined in Section~\ref{sec:query-adaptive-selection}. We keep representative-query sampling and budget-label computation unchanged and vary the regression scheme. On Llama-3.1-8B-Instruct, both local schemes achieve higher mean RULER scores than the global fit, supporting local estimation of query-dependent budgets. Sliding-window Linear Regression achieves the highest mean RULER score (Appendix~\ref{app:budget-regression}).

\paragraph{Number of shared rankings.}
We compare ProxyAttn and QASparse with 1, 2, and 4 shared rankings on GLM and Qwen, and with 1, 2, 4, and 8 on Llama and MegaBeam. With one shared ranking, QASparse achieves a higher mean RULER score than ProxyAttn on all four models. Reducing the count from the largest tested value to one also preserves accuracy better for QASparse on each model. QASparse therefore tolerates greater ranking sharing in these experiments, supporting the use of one shared ranking per layer. Appendix~\ref{app:shared-ranking} provides the per-length scores.

\paragraph{Boundary Voting candidate blocks.}
We vary the total number of candidate key blocks around the shared-ranking cutoff among 4, 8, 16, and 32, with Boundary Voting preserving the selected-block budget in each setting. The default 16-block region uses up to eight blocks on each side of the cutoff. On Llama-3.1-8B-Instruct, this setting achieves the highest mean RULER score, supporting the default candidate region (Appendix~\ref{app:boundary-voting}).

\section{Related Work}

\paragraph{Long Context LLMs.}
Existing research broadly follows three routes: continued pretraining or parameter-efficient fine-tuning on long text to adapt models to longer positional distributions\citep{data_engineering_128k_2024,longlora_2024,effective_long_context_2024,yangQwen251MTechnicalReport2025,deepseek_v4_2026}; modifying positional representations or context extrapolation schemes to enlarge the window without fully retraining the model\citep{pengYaRNEfficientContext2026,dingLongRoPE2024,longrope2_2025,gelbergExtendingContextPretrained2026,ms_poe_2024}; and introducing external memory, retrieval, or compression so that every decoding step need not attend to all historical tokens with equal intensity\citep{landmark_attention_2023,focused_transformer_2023,retrieval_long_context_2024,infini_attention_2024,longllmlingua_2024}. Longformer\citep{longformer_2020} and BigBird\citep{bigbird_2020} establish fixed sparse architectures based on local, global, and random connections.

\paragraph{LLM Inference Acceleration.}
Inference acceleration can be organized around exact-attention kernels, model-level sparsity, and KV-cache access. FlashAttention\citep{FLASHATTENTION:FastandMemory-EfficientExactAttentionwithIO-Awareness}, FlashAttention-2\citep{FlashAttention-2:FasterAttentionwithBetterParallelismandWorkPartitioning}, and FlashInfer\citep{yeFLASHINFEREFFICIENTCUSTOMIZABLE} optimize exact attention through IO-aware execution, parallel work partitioning, and serving-oriented kernels. Native Sparse Attention\citep{NativeSparseAttention:Hardware-AlignedandNativelyTrainableSparseAttention}, DeepSeek-V3.2\citep{DeepSeek-V3.2:PushingtheFrontierofOpenLargeLanguageModels}, and DeepSeek-V4\citep{deepseek_v4_2026} incorporate executable sparsity into training or model design. Prefill methods such as MInference\citep{jiangMInference10Accelerating}, FlexPrefill\citep{FlexPrefill:AContext-AwareSparseAttentionMechanismforEfficientLong-SequenceInference}, QuoKA\citep{jonesQUOKAQueryOrientedKV2026a}, VSPrefill\citep{vsprefill_2026}, and FlashPrefill V2\citep{flashprefill_v2_2026} select context-dependent token subsets, whereas decode methods reduce or selectively access the KV cache through H\textsuperscript{2}O\citep{h2o_2023}, StreamingLLM\citep{xiaoEfficientStreamingLanguage2024}, SnapKV\citep{liSnapKVLLMKnows}, Quest\citep{tangQuestQueryAwareSparsity2024}, DuoAttention\citep{xiaoDuoAttentionEfficientLongContext2025}, Tactic\citep{zhuTacticAdaptiveSparse}, Evolving Sparsity\citep{evolving_sparsity_2026}, and Louver\citep{louver_2026}. InfLLM-V2\citep{infllmv2_2026} connects these stages with switchable dense-sparse attention.

\section{Conclusion}
\label{sec:conclusion}

In this paper, we introduced QASparse, a query-adaptive sparse attention method for efficient long-context prefilling in LLMs. QASparse combines Local Budget Regression and Boundary Voting to adapt computation budgets and retained KV tokens to different query regions and attention heads while reusing shared token rankings. Length-Specialized Kernels further improve execution efficiency by tailoring the computation to short and long contexts. Experiments across four model families on RULER and InfiniteBench show that QASparse achieves strong model performance relative to existing sparse attention methods, while latency measurements demonstrate favorable accuracy--efficiency trade-offs. These results highlight the value of combining query-adaptive token selection with efficient kernel design for long-context inference. Future work could extend this design to other hardware platforms and further reduce the overhead of adaptive token selection.

\subsection*{AI use statement}
We used OpenAI Codex to assist with research ideation, method implementation, experiment design and execution, code debugging, result analysis, and manuscript writing. All AI-assisted content was reviewed and verified by the authors, who take full responsibility for the final work.

\bibliographystyle{iclr2027_conference}
\bibliography{main}

\clearpage
\appendix
\FloatBarrier
\section{Attention Ordering and Generation Quality}
\label{app:attention-importance}

We compare two token-selection orders during decoding using Llama-3.1-8B base on RULER at 4k. At each decoding step, we compute full attention and retain tokens in descending (\textit{high-to-low}) or ascending (\textit{low-to-high}) order of attention weight until their cumulative mass reaches 20\%, 30\%, or 40\%. All settings use the same 20 fixed examples from each of the 13 RULER tasks and full attention during prefill.

\begin{table}[H]
\centering
\small
\setlength{\tabcolsep}{3pt}
\renewcommand{\arraystretch}{1.08}
\caption{RULER scores of Llama-3.1-8B base at 4k, averaged equally over 13 tasks with 20 fixed examples per task.}
\label{tab:attention-ordering}
\begin{tabular*}{0.75\linewidth}{@{\extracolsep{\fill}}lrrr@{}}
\toprule
Threshold & High-to-low & Low-to-high & Gap \\
\midrule
20\% & 67.53 & 19.19 & +48.33 \\
30\% & 77.79 & 45.69 & +32.10 \\
40\% & 84.92 & 70.15 & +14.77 \\
\bottomrule
\end{tabular*}
\end{table}

Prioritizing high-attention tokens gives better generation quality at every tested coverage threshold, supporting attention weight as a criterion for token selection under a limited budget.

\FloatBarrier
\section{Number of Shared Rankings}
\label{app:shared-ranking}

Tables~\ref{tab:ranking-glm}--\ref{tab:ranking-qwen} report the RULER scores of ProxyAttn and QASparse on four models with different numbers of shared rankings. Each model is evaluated at six context lengths from 4k to 128k; Mean summarizes these lengths. Within each ranking-count pair, the higher score in each column is bold; ties highlight QASparse.

\begin{table}[H]
\centering
\small
\setlength{\tabcolsep}{3pt}
\renewcommand{\arraystretch}{1.08}
\caption{RULER scores of GLM-4-9B-Chat-1M across shared-ranking counts.}
\label{tab:ranking-glm}
\begin{tabular*}{\linewidth}{@{\extracolsep{\fill}}lrrrrrrrr@{}}
\toprule
Method & Rankings & 4k & 8k & 16k & 32k & 64k & 128k & Mean $\uparrow$ \\
\midrule
ProxyAttn & 1 & 95.00 & \textbf{93.67} & 91.73 & \textbf{90.57} & \textbf{87.76} & 85.45 & 90.70 \\
QASparse & 1 & \textbf{95.05} & 93.46 & \textbf{92.97} & 90.39 & 87.53 & \textbf{86.03} & \textbf{90.91} \\
\arrayrulecolor{black!20}\midrule[0.3pt]\arrayrulecolor{black}
ProxyAttn & 2 & \textbf{95.31} & 93.25 & 92.21 & 90.21 & \textbf{87.60} & 85.97 & 90.76 \\
QASparse & 2 & 95.29 & \textbf{93.77} & \textbf{93.31} & \textbf{91.24} & 86.05 & \textbf{86.31} & \textbf{90.99} \\
\arrayrulecolor{black!20}\midrule[0.3pt]\arrayrulecolor{black}
ProxyAttn & 4 & 95.21 & 93.73 & 92.44 & \textbf{90.62} & \textbf{87.70} & 85.61 & 90.89 \\
QASparse & 4 & \textbf{95.36} & \textbf{93.85} & \textbf{93.33} & 90.22 & 86.86 & \textbf{86.01} & \textbf{90.94} \\
\bottomrule
\end{tabular*}
\end{table}

\begin{table}[H]
\centering
\small
\setlength{\tabcolsep}{3pt}
\renewcommand{\arraystretch}{1.08}
\caption{RULER scores of Llama-3.1-8B-Instruct across shared-ranking counts.}
\label{tab:ranking-llama}
\begin{tabular*}{\linewidth}{@{\extracolsep{\fill}}lrrrrrrrr@{}}
\toprule
Method & Rankings & 4k & 8k & 16k & 32k & 64k & 128k & Mean $\uparrow$ \\
\midrule
ProxyAttn & 1 & 94.12 & 92.25 & 93.35 & 90.34 & 87.24 & 75.15 & 88.74 \\
QASparse & 1 & \textbf{94.90} & \textbf{93.21} & \textbf{93.75} & \textbf{92.22} & \textbf{87.70} & \textbf{77.82} & \textbf{89.93} \\
\arrayrulecolor{black!20}\midrule[0.3pt]\arrayrulecolor{black}
ProxyAttn & 2 & \textbf{94.68} & \textbf{92.55} & \textbf{93.39} & 90.17 & 86.52 & \textbf{77.60} & \textbf{89.15} \\
QASparse & 2 & 94.29 & 91.91 & 93.34 & \textbf{90.56} & \textbf{87.11} & 77.40 & 89.10 \\
\arrayrulecolor{black!20}\midrule[0.3pt]\arrayrulecolor{black}
ProxyAttn & 4 & 94.42 & \textbf{93.05} & 94.03 & 90.61 & 86.00 & \textbf{78.04} & 89.36 \\
QASparse & 4 & \textbf{94.82} & 92.73 & \textbf{94.28} & \textbf{91.73} & \textbf{86.48} & 77.35 & \textbf{89.56} \\
\arrayrulecolor{black!20}\midrule[0.3pt]\arrayrulecolor{black}
ProxyAttn & 8 & \textbf{95.28} & \textbf{93.21} & \textbf{94.39} & 90.71 & 86.41 & 76.68 & 89.44 \\
QASparse & 8 & 95.22 & 92.83 & 93.94 & \textbf{91.25} & \textbf{87.12} & \textbf{77.17} & \textbf{89.59} \\
\bottomrule
\end{tabular*}
\end{table}

\begin{table}[H]
\centering
\small
\setlength{\tabcolsep}{3pt}
\renewcommand{\arraystretch}{1.08}
\caption{RULER scores of MegaBeam-Mistral-7B-512k across shared-ranking counts.}
\label{tab:ranking-megabeam}
\begin{tabular*}{\linewidth}{@{\extracolsep{\fill}}lrrrrrrrr@{}}
\toprule
Method & Rankings & 4k & 8k & 16k & 32k & 64k & 128k & Mean $\uparrow$ \\
\midrule
ProxyAttn & 1 & 93.51 & 90.84 & \textbf{88.43} & \textbf{87.09} & \textbf{83.80} & 84.39 & 88.01 \\
QASparse & 1 & \textbf{94.21} & \textbf{92.30} & 87.96 & 86.94 & 83.00 & \textbf{85.30} & \textbf{88.29} \\
\arrayrulecolor{black!20}\midrule[0.3pt]\arrayrulecolor{black}
ProxyAttn & 2 & 93.13 & 90.32 & \textbf{86.74} & \textbf{86.72} & \textbf{83.98} & 84.82 & 87.62 \\
QASparse & 2 & \textbf{93.80} & \textbf{91.03} & 86.63 & 86.12 & 82.76 & \textbf{85.62} & \textbf{87.66} \\
\arrayrulecolor{black!20}\midrule[0.3pt]\arrayrulecolor{black}
ProxyAttn & 4 & 93.71 & 90.92 & \textbf{87.23} & \textbf{87.49} & \textbf{84.22} & \textbf{84.63} & 88.03 \\
QASparse & 4 & \textbf{94.09} & \textbf{91.76} & 87.02 & 87.25 & 83.42 & 84.62 & \textbf{88.03} \\
\arrayrulecolor{black!20}\midrule[0.3pt]\arrayrulecolor{black}
ProxyAttn & 8 & \textbf{93.67} & 91.53 & 88.11 & 87.17 & \textbf{84.13} & 85.44 & 88.34 \\
QASparse & 8 & 93.63 & \textbf{92.23} & \textbf{88.49} & \textbf{87.57} & 83.76 & \textbf{85.56} & \textbf{88.54} \\
\bottomrule
\end{tabular*}
\end{table}

\begin{table}[H]
\centering
\small
\setlength{\tabcolsep}{3pt}
\renewcommand{\arraystretch}{1.08}
\caption{RULER scores of Qwen2.5-7B-Instruct-1M across shared-ranking counts.}
\label{tab:ranking-qwen}
\begin{tabular*}{\linewidth}{@{\extracolsep{\fill}}lrrrrrrrr@{}}
\toprule
Method & Rankings & 4k & 8k & 16k & 32k & 64k & 128k & Mean $\uparrow$ \\
\midrule
ProxyAttn & 1 & 94.17 & 91.91 & 92.42 & 87.60 & 86.04 & 81.17 & 88.88 \\
QASparse & 1 & \textbf{94.35} & \textbf{92.21} & \textbf{92.91} & \textbf{88.95} & \textbf{86.54} & \textbf{82.52} & \textbf{89.58} \\
\arrayrulecolor{black!20}\midrule[0.3pt]\arrayrulecolor{black}
ProxyAttn & 2 & \textbf{94.63} & 91.58 & 92.30 & 87.74 & 86.82 & 81.27 & 89.06 \\
QASparse & 2 & 94.24 & \textbf{92.23} & \textbf{93.14} & \textbf{88.76} & \textbf{87.92} & \textbf{83.08} & \textbf{89.89} \\
\arrayrulecolor{black!20}\midrule[0.3pt]\arrayrulecolor{black}
ProxyAttn & 4 & 94.29 & \textbf{91.71} & 92.63 & 88.80 & 86.93 & 82.27 & 89.44 \\
QASparse & 4 & \textbf{94.51} & 91.54 & \textbf{92.87} & \textbf{89.31} & \textbf{87.84} & \textbf{83.73} & \textbf{89.97} \\
\bottomrule
\end{tabular*}
\end{table}

\FloatBarrier
\section{Boundary Voting Candidate-Block Count}
\label{app:boundary-voting}

Table~\ref{tab:boundary-voting-count} reports the candidate-block sweep discussed in Section~\ref{sec:ablations}, using Llama-3.1-8B-Instruct on RULER. The default 16-block setting takes up to eight blocks on each side of the shared-ranking cutoff, with a block traversal stride of one. Query and key vectors are sampled every eight positions within each 128-token block. \revised{These counts refer to total candidate blocks; $r$ in Section~\ref{sec:query-adaptive-selection} denotes the maximum number of candidate blocks on each side.} Scores are compared within each table; the component ablations and shared-ranking sweep use separate evaluation runs.

\begin{table}[H]
\centering
\small
\setlength{\tabcolsep}{3pt}
\renewcommand{\arraystretch}{1.08}
\caption{RULER scores of Llama-3.1-8B-Instruct for different total candidate-block counts.}
\label{tab:boundary-voting-count}
\begin{tabularx}{\linewidth}{@{}>{\raggedright\arraybackslash}Xrrrrrrr@{}}
\toprule
Setting & 4k & 8k & 16k & 32k & 64k & 128k & Mean $\uparrow$ \\
\midrule
4 candidate blocks & 92.67 & 92.29 & 91.69 & 88.90 & 88.73 & 80.41 & 89.12 \\
8 candidate blocks & 92.74 & 92.55 & 91.50 & 89.13 & 89.00 & 79.55 & 89.08 \\
16 candidate blocks & 92.67 & 93.51 & 91.76 & 89.67 & 88.92 & 78.71 & 89.21 \\
32 candidate blocks & 91.78 & 92.74 & 91.83 & 89.44 & 88.40 & 79.68 & 88.98 \\
\bottomrule
\end{tabularx}
\end{table}

\FloatBarrier
\section{Short- and Long-Context Kernel Timing}
\label{app:kernel-timing}

Table~\ref{tab:kernel-timing} gives the selection-kernel latencies underlying Figure~\ref{fig:kernel-speedup}, for seven context lengths from 8k to 512k. For short- and long-context latencies $t_s(L)$ and $t_l(L)$, the reference is their geometric mean, and each speedup is the reference divided by the corresponding latency:
\[
t_{\mathrm{ref}}(L)=\sqrt{t_s(L)t_l(L)},\qquad
S_s(L)=\frac{t_{\mathrm{ref}}(L)}{t_s(L)},\qquad
S_l(L)=\frac{t_{\mathrm{ref}}(L)}{t_l(L)}.
\]

\begin{table}[H]
\centering
\small
\setlength{\tabcolsep}{3pt}
\renewcommand{\arraystretch}{1.08}
\caption{Selection-kernel latency and speedup relative to the geometric-mean latency at each context length.}
\label{tab:kernel-timing}
\begin{tabular*}{\linewidth}{@{\extracolsep{\fill}}lrrrrr@{}}
\toprule
Length & \shortstack[r]{Short\\(ms)} & \shortstack[r]{Long\\(ms)} & \shortstack[r]{Reference\\(ms)} & \shortstack[r]{Short speedup\\($\times$)} & \shortstack[r]{Long speedup\\($\times$)} \\
\midrule
8k & 0.42 & 0.77 & 0.569 & 1.354 & 0.739 \\
16k & 1.16 & 1.50 & 1.319 & 1.137 & 0.879 \\
32k & 4.10 & 2.56 & 3.240 & 0.790 & 1.266 \\
64k & 78.28 & 5.31 & 20.388 & 0.260 & 3.840 \\
128k & 460.30 & 11.82 & 73.761 & 0.160 & 6.240 \\
256k & 1165.54 & 35.51 & 203.441 & 0.175 & 5.729 \\
512k & 6848.10 & 148.62 & 1008.843 & 0.147 & 6.788 \\
\bottomrule
\end{tabular*}
\end{table}

\FloatBarrier
\section{Local Budget Regression Scheme}
\label{app:budget-regression}

Table~\ref{tab:budget-regression} compares the three regression schemes defined in Section~\ref{sec:query-adaptive-selection} on Llama-3.1-8B-Instruct across the six RULER lengths. All schemes use one representative query per 128-token query block. Non-overlapping Piecewise Linear Regression uses 10 sampled queries per region. Sliding-window Linear Regression uses windows of 10 sampled queries, a stride of two sampled queries, and averages overlapping predictions.

\begin{table}[H]
\centering
\small
\setlength{\tabcolsep}{3pt}
\renewcommand{\arraystretch}{1.08}
\caption{RULER scores of Llama-3.1-8B-Instruct for the three budget-regression schemes.}
\label{tab:budget-regression}
\begin{tabularx}{\linewidth}{@{}>{\raggedright\arraybackslash}Xrrrrrrr@{}}
\toprule
Setting & 4k & 8k & 16k & 32k & 64k & 128k & Mean $\uparrow$ \\
\midrule
Global Linear Regression & 92.74 & 92.74 & 91.83 & 88.51 & 89.15 & 76.74 & 88.62 \\
Non-overlapping Piecewise Linear Regression & 92.74 & 92.74 & 91.06 & 89.74 & 89.04 & 79.68 & 89.17 \\
Sliding-window Linear Regression & 92.67 & 93.51 & 91.76 & 89.67 & 88.92 & 78.71 & 89.21 \\
\bottomrule
\end{tabularx}
\end{table}

\FloatBarrier
\section{Full Attention on the Final Two Query Blocks}
\label{app:final-query-blocks}

\begin{revision}
The final-mask rule gives the last two query blocks full attention over their causally visible keys. We evaluate both settings for FlexPrefill, ProxyAttn, and QASparse on Llama-3.1-8B-Instruct. In particular, the ProxyAttn baseline used in the enabled comparison also receives this rule.
\begin{table}[H]
\begin{revision}

\centering

\small
\setlength{\tabcolsep}{2pt}
\renewcommand{\arraystretch}{1.08}
\caption{\revised{RULER mean scores with and without full attention on the final two query blocks.}}
\label{tab:final-query-comparison}
\begin{tabular*}{\linewidth}{@{\extracolsep{\fill}}lrr@{}}
\toprule
Method & Enabled & Disabled \\
\midrule
FlexPrefill & 88.28 & 88.59 \\
ProxyAttn & 88.67 & 81.27 \\
QASparse & 89.21 & 84.69 \\
\bottomrule
\end{tabular*}

\end{revision}
\end{table}
\end{revision}
\begin{table}[H]
\centering
\small
\setlength{\tabcolsep}{3pt}
\renewcommand{\arraystretch}{1.08}
\caption{RULER scores with and without the final-two-query-block full-attention override.}
\label{tab:final-query-blocks}
\begin{tabularx}{\linewidth}{@{}>{\raggedright\arraybackslash}Xrrrrrrr@{}}
\toprule
Setting & 4k & 8k & 16k & 32k & 64k & 128k & Mean $\uparrow$ \\
\midrule
Force final two query blocks & 92.67 & 93.51 & 91.76 & 89.67 & 88.92 & 78.71 & 89.21 \\
No final-block override & 91.90 & 90.21 & 86.08 & 83.40 & 83.65 & 72.88 & 84.69 \\
\bottomrule
\end{tabularx}
\end{table}

\begin{revision}
Removing the rule reduces ProxyAttn by 7.40 points and QASparse by 4.52 points. QASparse exceeds ProxyAttn both with the rule (89.21 versus 88.67) and without it (84.69 versus 81.27). Its mean sparsity changes from 61.68\% without the rule to 59.73\% with it, a difference of 1.95 percentage points. These sparsity measurements accompany this final-block ablation.
\end{revision}
\FloatBarrier
\section{Ranking Used for Budget-Label Computation}
\label{app:budget-label-ranking}

This experiment changes the ordering used to construct the sampled budget labels in Local Budget Regression. Both variants compute the target head's actual post-RoPE attention mass and accumulate it toward the same coverage threshold. The default traverses key blocks in the proxy/shared ranking; the control traverses them in descending order of their true attention mass. Both variants use the shared ranking for initial selection, followed by the same Boundary Voting procedure, so only the ordering used to determine the budget labels is changed.

For normalized attention mass $a_j$ over causally visible key blocks, the label under ordering $\pi$ is
\[
b(\pi)=\min\left\{m:\sum_{j=1}^{m}a_{\pi(j)}\ge\gamma\right\}.
\]
Sorting by true mass reaches the threshold with no more blocks than the shared ordering. However, applying that budget to blocks selected in shared order may retain less mass than the label assumes. The comparison therefore tests whether budget estimation should follow the ordering used for selection. Table~\ref{tab:budget-label-ranking} reports both variants on Llama-3.1-8B-Instruct over the six RULER lengths, with the remaining settings unchanged.

\begin{table}[H]
\centering
\small
\setlength{\tabcolsep}{3pt}
\renewcommand{\arraystretch}{1.08}
\caption{RULER scores for shared and true orderings in budget-label computation; initial selection uses the shared ranking in both cases.}
\label{tab:budget-label-ranking}
\begin{tabularx}{\linewidth}{@{}>{\raggedright\arraybackslash}Xrrrrrrr@{}}
\toprule
Setting & 4k & 8k & 16k & 32k & 64k & 128k & Mean $\uparrow$ \\
\midrule
Shared ranking for budget label & 92.67 & 93.51 & 91.76 & 89.67 & 88.92 & 78.71 & 89.21 \\
True ranking for budget label & 91.90 & 92.63 & 90.83 & 89.17 & 86.85 & 76.46 & 87.97 \\
\bottomrule
\end{tabularx}
\end{table}

Using shared-order labels gives higher scores at every tested length, supporting alignment between budget estimation and the ordering used for selection. The experiment compares budget-label construction, rather than replacing the selection ranking with the true ranking.

\FloatBarrier
\section{Max Voting versus Exponential-Sum Voting}
\label{app:vote-scoring}

This experiment ablates the candidate-block scoring rule in Boundary Voting. Let $K_j$ contain the sampled key vectors of candidate block $j$, and let $\ell(q,k)$ denote the post-RoPE QK logit for sampled query $q$ and key $k$. Max voting scores a block by its strongest sampled interaction, whereas exponential-sum voting accumulates the exponentiated logits within the block:
\[
s_j^{\mathrm{max}}(q)=\max_{k\in K_j}\ell(q,k),\qquad
s_j^{\mathrm{exp}}(q)=\sum_{k\in K_j}\exp\!\bigl(\ell(q,k)\bigr).
\]
The exponential sum is computed with numerical stabilization, and only causally visible pairs are scored. In both variants, each sampled query casts exactly one vote for the candidate block with the highest score. Candidate blocks are then ranked by their vote counts, and the same number of blocks is retained. The candidate region, query/key sampling, and selected-block count are unchanged, so the ablation compares the evidence used to cast each vote rather than changing the voting budget.

Table~\ref{tab:voting-score} compares the two scoring rules on Llama-3.1-8B-Instruct across the six RULER lengths.

\begin{table}[H]
\centering
\small
\setlength{\tabcolsep}{3pt}
\renewcommand{\arraystretch}{1.08}
\caption{RULER scores for maximum and exponential-sum scoring in Boundary Voting.}
\label{tab:voting-score}
\begin{tabularx}{\linewidth}{@{}>{\raggedright\arraybackslash}Xrrrrrrr@{}}
\toprule
Setting & 4k & 8k & 16k & 32k & 64k & 128k & Mean $\uparrow$ \\
\midrule
Max voting & 92.67 & 93.51 & 91.76 & 89.67 & 88.92 & 78.71 & 89.21 \\
Exponential-sum voting & 91.97 & 92.74 & 91.83 & 89.67 & 89.46 & 78.91 & 89.10 \\
\bottomrule
\end{tabularx}
\end{table}

Max voting achieves the higher mean RULER score, supporting its use as the default voting rule.
\FloatBarrier

\FloatBarrier
\Needspace{0.72\textheight}
\begin{revision}
\section{Pareto Trade-off Measurements}
\label{app:pareto-data}
Figure~\ref{fig:pareto-all} compares RULER scores and attention latency for QASparse, ProxyAttn, and Delta Attention~\citep{willette2025delta} applied to StreamingLLM~\citep{xiaoEfficientStreamingLanguage2024} and HiP~\citep{leeHiPAttention2025} on Llama-3.1-8B-Instruct across six context lengths and their average.

\begin{figure}[H]
\begin{revision}

\centering
\includegraphics[width=\linewidth]{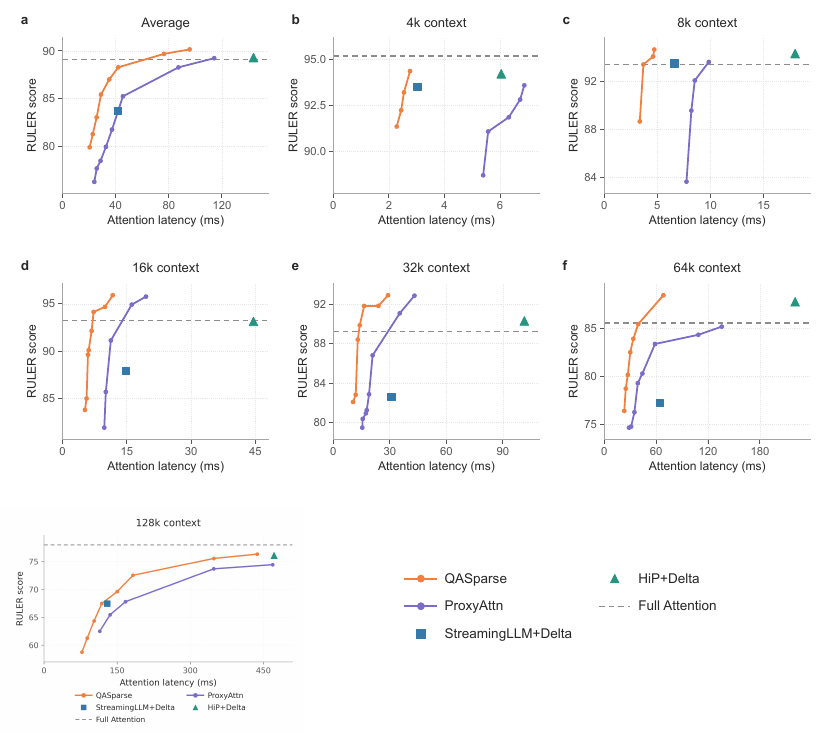}
\caption{\revised{RULER quality--latency trade-offs across context lengths.}}
\label{fig:pareto-all}

\end{revision}
\end{figure}
\end{revision}
\FloatBarrier
\Needspace{11\baselineskip}
\section{Attention Latency Measurements}
\label{app:attention-latency}

Table~\ref{tab:attention-latency-measurements} reports a separate attention-latency measurement of Llama-3.1-8B-Instruct at six context lengths, comparing FlashAttention with QASparse at $\gamma=0.90$.

\begin{table}[H]
\centering
\small
\setlength{\tabcolsep}{3pt}
\renewcommand{\arraystretch}{1.08}
\caption{Attention latencies (ms) of Llama-3.1-8B-Instruct.}
\label{tab:attention-latency-measurements}
\begin{tabular*}{\linewidth}{@{\extracolsep{\fill}}lrrrrrr@{}}
\toprule
\textbf{Method} & \textbf{4k} & \textbf{8k} & \textbf{16k} & \textbf{32k} & \textbf{64k} & \textbf{128k} \\
\midrule
FlashAttention & 0.71 & 2.48 & 9.47 & 36.57 & 143.77 & 595.57 \\
QASparse ($\gamma=0.90$) & 0.77 & 1.97 & 5.61 & 16.44 & 48.61 & 211.74 \\
\bottomrule
\end{tabular*}
\end{table}
\FloatBarrier

\Needspace{10\baselineskip}
\begin{revision}
\section{Selection-stage Memory Footprint}
\label{app:selection-memory}
Table~\ref{tab:selection-memory} reports the memory usage of each operation and its percentage of peak GPU memory for Llama-3.1-8B-Instruct at 64k and 128k, measured on an NVIDIA A800-SXM4-80GB GPU.

\begin{table}[H]
\begin{revision}

\centering
\fontsize{8}{10}\selectfont
\setlength{\tabcolsep}{1pt}
\renewcommand{\arraystretch}{1.13}
\caption{\revised{Memory usage in MB (percentage of peak GPU memory).}}
\label{tab:selection-memory}

\begin{tabular*}{\linewidth}{@{\extracolsep{\fill}}l*{7}{c}@{}}
\toprule
Context & Proxy Q/K & Proxy Scores & Sorting & Budget Labels & Regression & Voting & Attention \\
\midrule
64k & 141.62 (0.47\%) & 17.10 (0.06\%) & 3.12 (0.01\%) & 41.56 (0.14\%) & 0.09 (0.00\%) & 8.56 (0.03\%) & 538.37 (1.79\%) \\
128k & 275.39 (0.68\%) & 64.60 (0.16\%) & 11.76 (0.03\%) & 141.17 (0.35\%) & 0.17 (0.00\%) & 32.30 (0.08\%) & 1058.40 (2.60\%) \\
\bottomrule
\end{tabular*}

\end{revision}
\end{table}
\FloatBarrier
\end{revision}

\begin{revision}
\begingroup
\setlength{\intextsep}{4pt}
\FloatBarrier
\section{Component Ablation}
\label{app:component-isolation}

We ablate Local Budget Regression and Boundary Voting on all 13 RULER subtasks using Llama-3.1-8B-Instruct at $\gamma=0.95$. Table~\ref{tab:component-isolation} reports scores averaged across the 13 subtasks at each length.
\begin{table}[H]
\begin{revision}

\centering

\fontsize{8}{10}\selectfont
\setlength{\tabcolsep}{2pt}
\renewcommand{\arraystretch}{1.08}
\caption{\revised{Component ablations across all 13 RULER subtasks. Avg reports the six-length mean score.}}
\label{tab:component-isolation}
\begin{tabular*}{\linewidth}{@{\extracolsep{\fill}}lrrrrrrr@{}}
\toprule
Method & 4k & 8k & 16k & 32k & 64k & 128k & Avg \\
\midrule
ProxyAttn & 92.67 & 92.67 & 91.76 & 89.05 & 87.62 & 78.27 & 88.67 \\
Shared ranking + Boundary Voting & 92.67 & 92.67 & 92.72 & 89.05 & 87.92 & 78.08 & 88.85 \\
Shared ranking + Regression & 91.97 & 92.47 & 91.56 & 89.28 & 89.04 & 78.33 & 88.78 \\
QASparse & 92.67 & 93.51 & 91.76 & 89.67 & 88.92 & 78.71 & \textbf{89.21} \\
\bottomrule
\end{tabular*}

\end{revision}
\end{table}
Both modules improve the mean RULER score individually, and their combination achieves the highest mean score of 89.21, improving over ProxyAttn by 0.54 points.

\FloatBarrier
\section{Quality at a Coverage Threshold of 0.90}
\label{app:gamma90-quality}

Tables~\ref{tab:gamma90-ruler-llama}--\ref{tab:gamma90-infinite-glm} report the supplied quality measurements for $\gamma=0.90$ alongside $\gamma=0.95$ and the baseline methods. RULER means give equal weight to the six lengths. InfiniteBench means give equal weight to the ten listed tasks.
\begin{table}[H]
\begin{revision}

\centering

\fontsize{8}{10}\selectfont
\setlength{\tabcolsep}{2pt}
\renewcommand{\arraystretch}{1.08}
\caption{\revised{RULER scores of Llama-3.1-8B-Instruct.}}
\label{tab:gamma90-ruler-llama}
\begin{tabular*}{\linewidth}{@{\extracolsep{\fill}}lrrrrrrr@{}}
\toprule
Method & 4k & 8k & 16k & 32k & 64k & 128k & Mean \\
\midrule
FlashAttention-2 & 95.18 & 93.36 & 93.26 & 89.28 & 85.56 & 77.96 & 89.10 \\
FlexPrefill & 94.59 & 93.53 & 92.58 & 91.59 & 85.18 & 73.67 & 88.52 \\
ProxyAttn & 94.12 & 92.25 & 93.35 & 90.34 & 87.24 & 75.15 & 88.74 \\
QASparse ($\gamma=0.95$) & 94.90 & 93.21 & 93.75 & 92.22 & 87.70 & 77.82 & 89.93 \\
QASparse ($\gamma=0.90$) & 94.16 & 92.02 & 93.98 & 91.46 & 88.01 & 76.16 & 89.30 \\
\bottomrule
\end{tabular*}

\end{revision}
\end{table}
\begin{table}[H]
\begin{revision}

\centering

\fontsize{8}{10}\selectfont
\setlength{\tabcolsep}{2pt}
\renewcommand{\arraystretch}{1.08}
\caption{\revised{RULER scores of GLM-4-9B-Chat-1M.}}
\label{tab:gamma90-ruler-glm}
\begin{tabular*}{\linewidth}{@{\extracolsep{\fill}}lrrrrrrr@{}}
\toprule
Method & 4k & 8k & 16k & 32k & 64k & 128k & Mean \\
\midrule
FlashAttention-2 & 94.26 & 92.68 & 91.42 & 89.27 & 85.04 & 84.45 & 89.52 \\
FlexPrefill & 92.46 & 91.40 & 90.64 & 87.69 & 86.70 & 84.34 & 88.87 \\
ProxyAttn & 95.00 & 93.67 & 91.73 & 90.57 & 87.76 & 85.45 & 90.70 \\
QASparse ($\gamma=0.95$) & 95.05 & 93.46 & 92.97 & 90.39 & 87.53 & 86.03 & 90.91 \\
QASparse ($\gamma=0.90$) & 95.02 & 93.18 & 92.82 & 90.13 & 87.06 & 83.06 & 90.21 \\
\bottomrule
\end{tabular*}

\end{revision}
\end{table}
\begin{table}[H]
\begin{revision}

\centering

\fontsize{6.5}{8.5}\selectfont
\setlength{\tabcolsep}{2pt}
\renewcommand{\arraystretch}{1.08}
\caption{\revised{InfiniteBench scores of Llama-3.1-8B-Instruct. Task columns denote longbook summary, English longbook QA, longbook choice, dialogue QA, Chinese longbook QA, code execution, math finding, passkey, number-string retrieval, and KV retrieval, respectively.}}
\label{tab:gamma90-infinite-llama}
\begin{tabular*}{\linewidth}{@{\extracolsep{\fill}}lrrrrrrrrrrr@{}}
\toprule
Method & Sum. & QA En & Choice & Dial. & QA Zh & Code & Math & Passkey & Num. & KV & Mean \\
\midrule
FlashAttention-2 & 35.22 & 34.98 & 81.82 & 20.00 & 42.67 & 0.00 & 16.67 & 96.67 & 100.00 & 60.00 & 48.80 \\
FlexPrefill & 32.87 & 28.01 & 72.73 & 30.00 & 36.40 & 0.00 & 22.22 & 96.67 & 100.00 & 48.00 & 46.69 \\
ProxyAttn & 32.09 & 33.22 & 72.73 & 10.00 & 43.72 & 0.00 & 22.22 & 93.33 & 100.00 & 40.00 & 44.73 \\
QASparse ($\gamma=0.95$) & 35.13 & 38.58 & 72.73 & 10.00 & 41.42 & 0.00 & 22.22 & 96.67 & 100.00 & 56.00 & 47.28 \\
QASparse ($\gamma=0.90$) & 34.02 & 33.51 & 72.73 & 20.00 & 41.57 & 0.00 & 22.22 & 96.67 & 100.00 & 52.00 & 47.27 \\
\bottomrule
\end{tabular*}

\end{revision}
\end{table}
\begin{table}[H]
\begin{revision}

\centering

\fontsize{6.5}{8.5}\selectfont
\setlength{\tabcolsep}{2pt}
\renewcommand{\arraystretch}{1.08}
\caption{\revised{InfiniteBench scores of GLM-4-9B-Chat-1M. Task columns denote longbook summary, English longbook QA, longbook choice, dialogue QA, Chinese longbook QA, code execution, math finding, passkey, number-string retrieval, and KV retrieval, respectively.}}
\label{tab:gamma90-infinite-glm}
\begin{tabular*}{\linewidth}{@{\extracolsep{\fill}}lrrrrrrrrrrr@{}}
\toprule
Method & Sum. & QA En & Choice & Dial. & QA Zh & Code & Math & Passkey & Num. & KV & Mean \\
\midrule
FlashAttention-2 & 29.32 & 28.57 & 72.73 & 30.00 & 17.10 & 0.00 & 27.78 & 96.67 & 90.00 & 0.00 & 39.22 \\
FlexPrefill & 34.92 & 28.08 & 72.73 & 20.00 & 16.01 & 0.00 & 16.67 & 100.00 & 83.33 & 0.00 & 37.17 \\
ProxyAttn & 31.13 & 27.49 & 72.73 & 20.00 & 17.62 & 0.00 & 27.78 & 96.67 & 80.00 & 0.00 & 37.34 \\
QASparse ($\gamma=0.95$) & 28.78 & 28.47 & 72.73 & 30.00 & 17.52 & 0.00 & 22.22 & 96.67 & 80.00 & 0.00 & 37.64 \\
QASparse ($\gamma=0.90$) & 29.23 & 25.48 & 72.73 & 30.00 & 17.26 & 0.00 & 22.22 & 96.67 & 73.33 & 0.00 & 36.69 \\
\bottomrule
\end{tabular*}

\end{revision}
\end{table}
On Llama RULER, QASparse at $\gamma=0.90$ scores 89.30, above the other listed methods except QASparse at $0.95$ (89.93). On GLM, the corresponding scores are 90.21 and 90.91; ProxyAttn scores 90.70. On InfiniteBench, Llama's mean is 47.27 at $0.90$ and 47.28 at $0.95$ (a 0.01-point difference), while GLM changes from 37.64 to 36.69 (0.95 points).

\FloatBarrier
\section{System Measurements}
\label{app:kkl3-system}

\paragraph{Peak memory.} We compare QASparse at $\gamma=0.90$ with FlashAttention-2 at 64k and 128k. Relative increases are computed as $(M_{\mathrm{QASparse}}/M_{\mathrm{FA2}}-1)\times100\%$ using the recorded GiB values.
\begin{table}[H]
\begin{revision}

\centering

\fontsize{8}{10}\selectfont
\setlength{\tabcolsep}{2pt}
\renewcommand{\arraystretch}{1.08}
\caption{\revised{Peak GPU memory (GiB) and relative increase over FlashAttention-2.}}
\label{tab:kkl3-peak-memory}
\begin{tabular*}{\linewidth}{@{\extracolsep{\fill}}lrrrr@{}}
\toprule
Model & Length & FA2 & QASparse & Increase \\
\midrule
Llama-3.1-8B-Instruct & 64k & 28.12 & 28.25 & 0.46\% \\
Llama-3.1-8B-Instruct & 128k & 38.66 & 39.15 & 1.27\% \\
MegaBeam-Mistral-7B-512k & 64k & 26.65 & 26.78 & 0.49\% \\
MegaBeam-Mistral-7B-512k & 128k & 37.19 & 37.68 & 1.32\% \\
GLM-4-9B-Chat-1M & 64k & 29.27 & 29.40 & 0.44\% \\
GLM-4-9B-Chat-1M & 128k & 38.05 & 38.55 & 1.31\% \\
Qwen2.5-7B-Instruct-1M & 64k & 23.39 & 23.50 & 0.47\% \\
Qwen2.5-7B-Instruct-1M & 128k & 29.11 & 29.56 & 1.55\% \\
\bottomrule
\end{tabular*}

\end{revision}
\end{table}
The relative increase is 0.44--0.49\% at 64k and 1.27--1.55\% at 128k across the four models.
\paragraph{Stage-time breakdown.} We profile QASparse on Llama-3.1-8B-Instruct at $\gamma=0.95$ over 128k--768k and report mean per-layer stage-time shares in Table~\ref{tab:kkl3-stages}. Sparse attention accounts for 94.68--98.14\% of the profiled stage time.
\begin{table}[H]
\begin{revision}

\centering

\fontsize{7.5}{9.5}\selectfont
\setlength{\tabcolsep}{2pt}
\renewcommand{\arraystretch}{1.08}
\caption{\revised{Mean per-layer stage-time shares (\%) for Llama-3.1-8B-Instruct at $\gamma=0.95$.}}
\label{tab:kkl3-stages}
\begin{tabular*}{\linewidth}{@{\extracolsep{\fill}}lrrrrrrr@{}}
\toprule
Context & Proxy Q/K & Proxy Scores & Sorting & Budget Labels & Regression & Voting & Attention \\
\midrule
128k & 0.85 & 0.89 & 0.03 & 2.98 & 0.03 & 0.54 & 94.68 \\
256k & 0.49 & 0.63 & 0.02 & 1.72 & 0.01 & 0.47 & 96.66 \\
512k & 0.03 & 0.51 & 0.01 & 1.75 & 0.01 & 0.20 & 97.49 \\
768k & 0.05 & 0.46 & 0.03 & 1.09 & 0.01 & 0.23 & 98.14 \\
\bottomrule
\end{tabular*}

\end{revision}
\end{table}

\FloatBarrier
\endgroup
\end{revision}

\begin{revision}
\Needspace{12\baselineskip}
\section{Illustrative Budget-estimation Errors}
\label{app:budget-error}

On selected RULER 4k examples, we measure budget-estimation error against the budget required for $\gamma=0.90$ attention coverage along the shared ranking. For query block $b$, let $w_b$ be its number of valid queries, and let $\bar{k}_{m,b}$ and $\bar{k}_{\mathrm{ref},b}$ denote the method's and reference's mean budgets in visible key tokens. We use
\begin{equation}
\mathrm{MAE}_m = \frac{\sum_b w_b\left|\bar{k}_{m,b}-\bar{k}_{\mathrm{ref},b}\right|}{\sum_b w_b}.
\end{equation}

\begin{figure}[H]
\begin{revision}

\centering
\includegraphics[width=\linewidth]{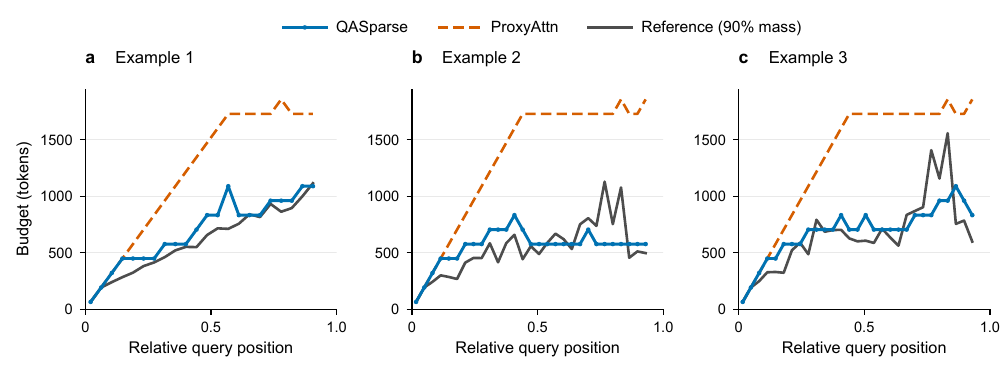}
\caption{\revised{Budget estimates on selected RULER 4k examples. Each point is a mean over valid queries in a 128-token query block, and horizontal positions are block centers divided by input length.}}
\label{fig:budget-error}

\end{revision}
\end{figure}

For Examples 1--3, QASparse's MAEs are 86.51, 144.01, and 144.35 tokens, compared with 601.33, 800.36, and 675.18 tokens for ProxyAttn. QASparse's MAE is 14.39\%, 17.99\%, and 21.38\% of ProxyAttn's, respectively.
\end{revision}

\end{document}